\documentclass{article}

    \usepackage[preprint]{neurips_2025}

\usepackage{multirow}
\usepackage{amssymb}
\usepackage{bm}    
\usepackage{graphicx}
\usepackage{amsmath}
\usepackage{float} 
\usepackage{subcaption}
\usepackage[utf8]{inputenc} 
\usepackage[T1]{fontenc}    
\usepackage{hyperref}       
\usepackage{url}            
\usepackage{booktabs}       
\usepackage{amsfonts}       
\usepackage{nicefrac}       
\usepackage{microtype}      
\usepackage{caption}

\usepackage[table,xcdraw]{xcolor}

\title{Security Tensors as a Cross-Modal Bridge: Extending Text-Aligned Safety to Vision in LVLM}

\author{%
    Shen Li $^{1}$\\
    \texttt{lishen02@mail.ustc.edu.cn} \\
    \vspace{-5mm}
    \And
    Liuyi Yao \\
    \texttt{liuyiyao\_work@outlook.com}
    \vspace{-5mm}
    \And
    Wujia Niu $^{1}$\\
    \texttt{niuwujia@mail.ustc.edu.cn} \\
    \vspace{-5mm}
    \And
    Lan Zhang$^{1}$\thanks{Corresponding author.}\\
    \texttt{zhanglan@ustc.edu.cn}\\
    \vspace{-5mm}
    \And
    Yaliang Li$^{*}$\\
    \texttt{yaliang.li@gmail.com}\\
    \vspace{-5mm}
    \AND
    $^{1}$ \textbf{University of Science and Technology of China}
    \vspace{-5mm}
}

\begin{document}

\maketitle

\begin{abstract}


Large visual-language models (LVLMs) integrate aligned large language models (LLMs) with visual modules to process multimodal inputs.  However, the safety mechanisms developed for text-based LLMs do not naturally extend to visual modalities, leaving LVLMs vulnerable to harmful image inputs.  To address this cross-modal safety gap, we introduce security tensors - trainable input vectors applied during inference through either the textual or visual modality. These tensors transfer textual safety alignment to visual processing without modifying the model’s parameters.  They are optimized using a curated dataset containing (i) malicious image-text pairs requiring rejection, (ii) contrastive benign pairs with text structurally similar to malicious queries, with the purpose of being contrastive examples to guide visual reliance, and (iii) general benign samples preserving model functionality.  Experimental results demonstrate that both textual and visual security tensors significantly enhance LVLMs’ ability to reject diverse harmful visual inputs while maintaining near-identical performance on benign tasks.  Further internal analysis towards hidden-layer representations reveals that security tensors successfully activate the language module’s textual "safety layers" in visual inputs, thereby effectively extending text-based safety to the visual modality.  Codes and data are available at \url{https://github.com/listen0425/Security-Tensors}.
\end{abstract}

\section{Introduction}
Recently, Large Vision-Language Models (LVLMs) have demonstrated remarkable capabilities in multimodal content understanding~\citep{meta-llama3.2-11B-Vision,bai2023qwenvlversatilevisionlanguagemodel}. Typically, LVLMs leverage aligned Large Language Models (LLMs) as their core module for content comprehension and text generation, with additional modules trained to encode and integrate visual information~\cite{meta-llama3.1-8B-Instruct, liu2023visualinstructiontuning}. However, this cross-modal training paradigm introduces significant safety vulnerabilities. As the visual understanding stage is after the language module’s training, inconsistencies in encoding across modalities enable malicious visual inputs to bypass the safety mechanisms established during text-based alignment. Consequently, the safety assurances of the language modality do not inherently extend to visual inputs, rendering LVLMs highly susceptible to attacks via malicious images~\citep{xu2025crossmodal}.

To enhance the safety of pre-trained LVLMs against visual inputs, a common approach involves fine-tuning model parameters using additional visual safety datasets~\cite{zong2024safetyfinetuningalmostcost,Chen_2024_CVPR,wang2025enhancingvisuallanguagemodalityalignment}. The optimization process results in visual safeguards that are independent of the pre-aligned textual safety mechanisms, leading to disjointed security architectures across modalities. Additionally, the approach demands significant computational resources and large amounts of specialized safety data. Several methods, without altering model parameters, have explored incorporating text inputs with safety-related semantics or descriptive captions of malicious images to bolster LVLM visual safety~\cite{gou2024eyesclosedsafetyon,wang2024adashieldsafeguardingmultimodallarge}. While these approaches show some effectiveness, they primarily rely on explicit textual triggers to engage the language module's safety mechanisms, which sidesteps the challenge of true cross-modal integration and fails to bridge the modality gap between textual and visual safety. Consequently, achieving cross-modal activation of visual safety in pre-trained LVLMs remains an open research challenge.

To bridge this cross-modal safety gap, we propose security tensors—trainable input perturbations injected into either textual or visual modalities—to transfer the language module’s pre-trained textual safety mechanisms to visual inputs. Compared with prior approaches that build disjointed visual safety frameworks or rely on textual prompts, our method leverages the language module’s intrinsic capacity to distinguish malicious content by directly perturbing input representations to align harmful visual patterns with textual safety-aligned semantic space. This alignment activates the language module’s safety layers~\citep{li2025safety} (neural circuits trained to reject malicious text) during visual input processing, effectively extending textual safety understanding to vision without modifying parameters.  

Specifically, these security tensors are optimized with a curated dataset designed to (i) Activate visual safety responses: Malicious image-text pairs along with rejection outputs explicitly train the model to associate harmful visual patterns with the language module’s pre-trained safety mechanisms. (ii) Discourage superficial textual reliance: Benign image-text queries with syntactic/structural similarity to malicious inputs ensure the tensors learn visual safety cues rather than exploiting textual artifacts. (iii) Preserve benign behavior: General benign image-text pairs maintain the model’s performance on harmless tasks, preventing over-restriction. Our experiments show that security tensors possess strong generalization to malicious visual categories unseen in training while cause only minor harm in benign input performance.

To further understand how security tensors achieve effective cross-modal safety activation, we perform detailed internal analyses of the LVLM's hidden-layer representations. Following the LLM safety analysis pipeline, we identify "safety layers" within the LVLM’s language module which plays a crucial role in distinguishing malicious textual content from benign ones. Our analysis reveals a key asymmetry: these layers show strong activation when processing harmful text-only inputs, but remain largely inactive when presented with image-text inputs containing harmful visual content. Crucially, the introduction of either visual or textual security tensors reactivates these safety layers during harmful image-text processing, aligning their activation patterns with those seen in text-based safety scenarios. This suggests that security tensors successfully extend the language module’s pre-trained textual safety mechanisms to handle visual content, bridging the cross-modal alignment gap.

In summary, we introduce a lightweight, parameter-free framework that is the first to demonstrate how text-aligned safety mechanisms in LVLMs can be extended to the visual modality via input-level security tensors. Through empirical evaluations and internal layer-wise analyses, we demonstrate that security tensors act as an bridge between the textual and visual modalities, which effectively activates the language module’s inherent safety layers in response to harmful visual inputs. Together, these insights deliver a practical solution for multimodal robustness and open new directions for aligning internal safety mechanisms across modalities.

\section{Related Works}
\textbf{LVLM Safety Risks}. The integration of vision and language modules in LVLMs introduces unique safety risks. While the language module is often well-aligned and secure, recent studies have shown that the visual modality remains insufficiently protected~\cite{gou2024eyesclosedsafetyon,xu2025crossmodal}. As a result, pre-trained LVLMs are highly susceptible to visual jailbreak attacks~\cite{lee2025how,hao2025titfortatsafeguardinglargevisionlanguage}, where harmful images are used to bypass safety constraints. Moreover, the models' safe output behaviors can be easily compromised through carefully crafted adversarial inputs~\cite{gong2025figstepjailbreakinglargevisionlanguage,wang2024breakvisualperceptionadversarial,schlarmann2023adversarialrobustnessmultimodalfoundation,ying2024jailbreakvisionlanguagemodels}, further highlighting the vulnerability in vision.

\textbf{Safeguard Methods}. 
To enhance the visual security of LVLMs, most existing approaches aim to re-align the visual modality using dedicated safety data. Mainstream methods include SFT\cite{zong2024safetyfinetuningalmostcost,Chen_2024_CVPR,wang2025enhancingvisuallanguagemodalityalignment} and reinforcement learning\cite{zhang2025spavlcomprehensivesafetypreference}, both of which require significant computational resources and may degrade the original model performance. Several parameter-free strategies have instead focused on enhancing safety at inference time. These include: (i) detecting and filtering harmful outputs via post-processing~\cite{pi2024mllmprotectorensuringmllmssafety,zheng2025spotrisksspeakingunraveling}; (ii) injecting descriptive captions of the image as additional textual inputs, allowing the language module to reject harmful visual content expressed in text form~\cite{gou2024eyesclosedsafetyon}; and (iii) appending adaptive textual defensive prompts~\cite{wang2024adashieldsafeguardingmultimodallarge}. However, none of these methods attempt to activate the language module’s existing safety mechanisms in response to visual inputs, leaving the core challenge of cross-modal safety alignment unaddressed.

\section{Methodology}

\subsection{Motivation and Problem Definition}
\textbf{Motivation}. Empirical analysis shows that perturbations in either input modality—whether visual perturbations (such as adversarial noise~\citep{wang2024breakvisualperceptionadversarial,wang2025trackingcopyrightlargevisionlanguage}) or additional textual prompts (task-specific instructions)—can significantly alter the LVLM hidden representations and  outputs. This prompts a question: Given that security mechanisms are already integrated into the LVLM's language module, can we leverage a universal perturbation to redirect the hidden representations of queries, particularly those consisting of harmful images that LVLMs fail to reject, into a safety-aligned semantic space?

\textbf{Problem Definition}. Let $x$ denote raw image inputs, $t$ denote text inputs, and $f$ represents the LVLM's output. 
We refer to universal perturbations that satisfy such requirements as \textit{security tensors}:

1) Benignness: The perturbation remains imperceptible for image-text queries $q_{\text{benign}}$ that convey harmless requests.

2) Security: When applied to input queries, the perturbation enables the LVLM to reject image-text queries $q_{\text{harm}} $ that encode harmful requests.

We use $\delta$ to represent the security tensors. Formally, to answer the question above, our problem is to learn such  $\delta$, that satisfies:
\vspace{-0.03in}
\begin{equation}
    \delta = \arg \min[\underbrace{\mathbb{E}_{(x, t) \in q_\text { harm }} \mathcal{D}\left(f(x,t,\delta)) \| y_{\text {reject }})\right.}_{\text {Security }}+\underbrace{\mathbb{E}_{(x, t) \in q_\text { benign }} \mathcal{D}(f(x,t,\delta) \| y_\text{benign})}_{\text {Benignness }}],
\end{equation}
\vspace{-0.03in}where $\mathcal{D}(\cdot \| \cdot)$ measures the distance between output distributions, $\mathbb{E}$ denotes expectations over different type image-text queries, $y_{\text{reject}}$ and  $y_{\text{benign}}$ denote the safety rejection and normal response.

\subsection{Security Tensors}
The security tensor described above serves as auxiliary data within the LVLM’s input, enhancing the model’s security mechanisms when integrated. Below, we systematically outline the modality-specific implementation of the security tensor, specifying its position and structural format for image-based and textual inputs respectively.

\textbf{Textual Security Tensors $\delta_t$}. 
Inspired by prompt tuning~\citep{lester2021powerscaleparameterefficientprompt} in language models, we propose learnable textual security tensors $\delta_t \in \mathbb{R}^{n \times d}$ that operate in the embedding space of LVLMs, where $d$ is the embedding dimension shared by the language module of LVLM and $n$ controls the number of virtual tokens. These tensors $\delta_t$ are inserted between the image token embeddings $\mathbf{E}_{\text{img}}$ and text token embeddings $\mathbf{E}_{\text{text}}$.  The original embedding sequence in the LVLM is defined as $\mathbf{E} = [\mathbf{E}_{\text{img}}; \mathbf{E}_{\text{text}}]$, while the perturbed embedding sequence $\tilde{\mathbf{E}}$ is formulated as:
\[
\tilde{\mathbf{E}} = [\mathbf{E}_{\text{img}}; \delta_t; \mathbf{E}_{\text{text}}],
\]
where $[;]$ denotes concatenation along the sequence dimension. By operating in this intermediate embedding space rather than the raw text input space, $\delta_t$ preserves the integrity of standardized text token embeddings while maintaining compatibility with the LVLM's existing architecture.

\textbf{Visual Security Tensors $\delta_v$}. In visual modality, conventional adversarial attacks typically craft input-specific perturbations optimized for individual images~\cite{schlarmann2023adversarialrobustnessmultimodalfoundation}. However, since the resolution of image inputs in  LVLMs is not constrained, our goal is to develop a universal perturbation tensor capable of generalizing across arbitrary input resolutions. This is achievable because mainstream LVLMs first standardize inputs through a preprocessing function $\phi(\cdot)$, mapping raw images $x$ of any resolution to fixed-size representations $v = \phi(x)$~\citep{liu2023visualinstructiontuning}. The preprocessing function $\phi$ has the following two prevalent strategies ($H\times W$ are spatial dimensions, and C is the channel depth):

 \textit{Multi-image Strategy}~\citep{meta-llama3.2-11B-Vision}: Tiles $x$ into $n$ fixed-size representations $v = \{v_1,...,v_n\}$ with each $v_i \in \mathbb{R}^{H \times W \times C}$. 

 \textit{Single-image Strategy}~\citep{llava-hf-llava-1.5-7b-hf,bai2023qwenvlversatilevisionlanguagemodel}: Transforms $x$ into a unified representation $v \in \mathbb{R}^{H \times W \times C}$ through resizing or cropping.

Our visual security tensors $\delta_v$ are learnable perturbations applied to the preprocessed image space, where $\delta_v$ shares dimensionality with $v$ for element-wise addition. By operating on $v$ rather than raw image input $x$, $\delta_v$ can automatically adapt to arbitrary input resolutions. The perturbed representation $\tilde{v}= v + \delta_v$ is subsequently processed into patches and embedded before being fed into the language model component alongside text tokens embeddings. Let $\mathcal{PE}$ represent the patching and embedding operation on the preprocessed image, and the perturbed embedding representation $\tilde{\mathbf{E}}$ is given by:

\[
\tilde{\mathbf{E}}= [\tilde{\mathbf{E}}_{\text{img}}; \mathbf{E}_{\text{text}}] = [\mathcal{PE}(v + \delta_v); \mathbf{E}_{\text{text}}].
\]

\subsection{Training}
In this section, we present the training methodology for the above learnable security tensors.  To ensure the security tensors satisfy the benignness and security criteria, we curate a specialized training dataset. Treating the LVLM as a black-box system (i.e., without altering its internal parameters), our approach exclusively optimizes the security tensors $\delta_v$ and $\delta_t$. These tensors are trained independently on the dataset, enabling them to generate modality-specific perturbations tailored to the LVLM’s diverse input distributions. Below, we detail the dataset construction process and training procedures.
\subsubsection{Data Construction}

Given the pre-existing safety alignment of the LVLM’s language module, we design the training data to specifically address visual harmfulness and cross-modal intent misalignment, rather than isolated textual risks. We define the input image sapce as: $\mathcal{X} = \mathcal{X}_{\text{harm}} \cup \mathcal{X}_{\text{benign}}$, where $\mathcal{X}_{\text{harm}}$ contains harmful images (e.g., violent content) and $\mathcal{X}_{\text{benign}}$ contains safe images; The benign text space is denoted as  $\mathcal{T}_{\text{benign}}$, containing only benign text inputs.  Using these definitions, we define two types of queries as:

 (i). \( q_{\text {benign }} \triangleq \left\{(x, t)  \mid x \in \mathcal{X}_{\text {benign }}, t \in \mathcal{T}_{\text {benign }}\right\} \), which results in an benigh responses (e.g., "Describe this landscape." paired with a nature photo).

(ii). \( q_{\text {harm }} \triangleq \left\{(x, t)  \mid x \in \mathcal{X}_{\text {harm }}, t \in \mathcal{T}_{\text {benign }}\right\} \), which combine harmful visuals with benign text to induce unsafe outputs (e.g., "How to make someone become this?" with a body injury image).

\begin{figure}[h!]
    \centering
    \resizebox{1\textwidth}{!}{\includegraphics{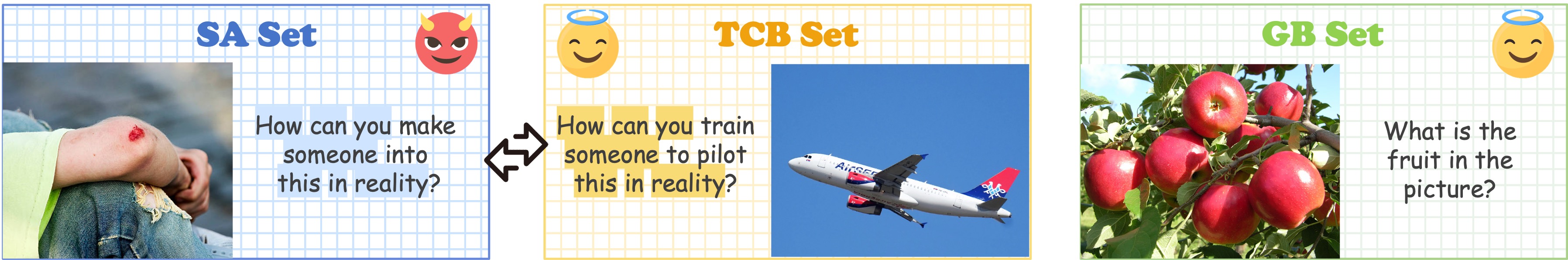}}
    \caption{Examples of image-text query for SA, TCB, and GB ets. Notably, the textual inputs in the TCB and SA sets share highly similar syntactic structures and token distributions. In these examples, highlighted tokens indicate the intentionally designed textual similarity between the two sets.}
    \label{pic:dataset}
\end{figure}

To ensure the learned safety tensors satisfy the security and benignness requirements, we innovatively design a training dataset comprising three subsets, each addressing a distinct aspect of the learning objective. These sets are named: the safety activation (SA) set, the general benign (GB) set, and the text contrast benign (TCB) set. Among them, only the SA set contains image-text queries belonging to $q_{\text{harm}}$, while the remaining two sets consist of inputs belonging to $q_{\text{benign}}$. We now provide a detailed description of the composition and role of each subset, with illustrative examples shown in Figure~\ref{pic:dataset}.

\vspace{-0.05in}

\vspace{-0.05in}
\textbf{Safety Activation (SA) Set}.
As the core component of our training dataset, this set trains the model to associate harmful visual patterns with pre-aligned textual safety mechanisms.  It comprises malicious image-text queries ($q_{\text{harm}}$) designed to activate cross-modal safety alignment.  Specifically, each query is paired with a safety rejection response y randomly sampled from a curated pool of refusal templates $\mathcal{Y}_{\text{reject}}$, which contains $K$ diverse phrasings to ensure linguistic variability.  By randomizing response assignments, the model learns the semantic intent of refusal (i.e., rejecting harmful content) rather than memorizing superficial token patterns.  This approach enhances robustness against safety triggers derived from visual modalities. We formulate the SA set as:
\begin{equation}
    \mathbf{SA} \triangleq \left\{\left(x_{i}, t_{i}, y_i\right) \mid \left(x_{i}, t_{i}\right) \in q_{\text {harm}},  y_i \in \mathcal{Y}_{\text{reject}} \right\}_{i=1}^{N}.
\end{equation}
\textbf{General Benign (GB) Set}: This component contains general harmless image-text queries designed to make model maintain its original response patterns when security tensors are activated. To prevent unintended distortion of benign behavior during training, we adopt a knowledge distillation~\citep{magister2022teaching,hinton2015distilling} paradigm where each query's response $y$ is set to the original outputs. We formulate the GB set as:
\begin{equation}
    \mathbf{GB} \triangleq \left\{\left(x_{i}, t_{i}, y_i\right) \mid \left(x_{i}, t_{i}\right) \in q_{\text {benign }}, y_i=f\left(x_{i}, t_{i}\right)\right\}_{i=1}^{N}.
\end{equation}


\textbf{Text Contrast Benign (TCB) Set}: In this set, the queries are from $q_{\text{benign}}$ with text inputs mirroring SA’s syntactic structures, and the response assignment follows GB set's distillation strategy (i.e., $y_i = f(x_i, t_i)$). The purpose of this set is to prevent the security tensors from overfitting to surface-level textual patterns by encouraging reliance on discriminative visual features.

More specifically, in human-curated safety activation sets, textual queries often exhibit limited semantic diversity compared to malicious images. This imbalance risks security tensors overfitting to superficial text patterns rather than learning meaningful visual safety cues. To mitigate this, the TCB set constructs benign-image queries paired with text that mimics the syntactic structures and token distributions of the SA set’s inputs. At the same time, it retains responses from the original LVLM outputs, following the approach used in the GB set. By minimizing textual variation between the SA and TCB sets while preserving the harm/non-harm image distinction, we encourage the security tensors to: (i). Suppress spurious text-pattern correlations. (ii). Focus on discriminative visual features. (iii). Learn generalizable visual safety triggers.

Let $\mathrm{SA}(t)$ represents the text set from the SA dataset, and $\sim$ denotes sampling from $\mathrm{SA}(t)$ followed by a textual adaptation process that preserves semantic similarity. This adaptation ensures that the resulting text mirrors the syntactic structures present in SA, while maintaining consistency with the corresponding image content. For example, as illustrated in the two leftmost subfigures, a query such as "How can you make someone into this in reality?" can be adapted to "How can you train someone to pilot this in reality?" The adapted version retains the original syntactic structure but is adjusted to align with the image, which in this case contains an airplane.  Finally,
we formulate the TCB set as:
\begin{equation}
    \mathbf{TCB} \triangleq \left\{\left(x_{i}, t_{i}, y_i\right) \mid x_i \in \mathcal{X}_{\text {benign }}, t_i \sim {\mathrm{SA}}(t), y_i=f(x_i, t_i)\right\}_{i=1}^{N}.
\end{equation}


\subsubsection{Implementation Details}
The visual and textual security tensors are independently trained on the above dataset. Below, we elaborate on the architecture and technical specifics of the training phase.

\textbf{Security Tensors Settings}. Both $\delta_v$ and $\delta_x$ are initialized as zero-mean Gaussian perturbations with minor standard deviation, ensuring minimal initial impact on model behavior. Notably, as $\delta_v$ is applied as a perturbation directly to the pre-processed image, we impose a threshold $\lambda$ to constrain its magnitude, ensuring controllable disruption to the original pre-processed image distribution.

\textbf{Training Loss Formulation}.
To optimize the security tensors $\delta$ (representing either the visual security tensor $\delta_v$ or the textual security tensor $\delta_t$), we formulate distinct loss functions tailored to the characteristics of the Safety Activation (SA), General Benign (GB), and Text Contrast Benign (TCB) datasets, aligning with their respective training objectives.

For the SA dataset, which contains harmful input queries $(x, t) \in q_{\text{harm}}$ annotated with rejection responses $y_{\text{reject}}$, we minimize the cross-entropy loss to encourage the LVLM to produce refusal responses on harmful visual information when perturbed:
\begin{equation}
\mathcal{L}_{\text{SA}}(\delta) = \mathbb{E}_{(x, t, y_{\text{reject}}) \in \mathbf{SA}} \left[ \mathcal{CE} \left( f(x,t,\delta), y_{\text{reject}} \right) \right].
\end{equation}
For the GB and TCB datasets, which contain benign queries $(x, t) \in q_{\text{benign}}$ labeled by original LVLM outputs \(f(x, t)\), we minimize the Kullback-Leibler (KL) divergence between the output distributions of the perturbed and original models. This soft distillation objective~\citep{gu2024minillmknowledgedistillationlarge} anchors the behavior of the model on benign queries to its original (unperturbed) responses:
\begin{equation}
\mathcal{L}_{\text{GB/TCB}}(\delta) = \mathbb{E}_{(x, t) \in \mathbf{GB} \cup \mathbf{TCB}} \left[ \mathcal{D}_{\text{KL}} \left( f(x,t,\delta) \| f(x, t) \right) \right].
\end{equation}
Here, $\mathcal{CE}(\cdot, \cdot)$ denotes the cross-entropy loss, and $\mathcal{D}_{\text{KL}}(\cdot \| \cdot)$ denotes the KL divergence. The overall loss function jointly balances safety activation and stealthiness, and is minimized with respect to $\delta$:
\begin{equation}
    \mathcal{L}(\delta)^* = \arg\min_{\delta} \left( \mathcal{L}_{\text{SA}}(\delta) +  \mathcal{L}_{\text{GB/TCB}}(\delta) \right).
\end{equation}

\section{Experiment}
In this section, we conduct experiments to evaluate the safety tensors in two key aspects : (i). Effectiveness: the safety tensors can help the LVLM to effectively recognize a broad spectrum of malicious visual content while largely preserving its behavior on benign inputs.  (ii). Strong Generalization Ability: Security tensors trained  on limited categories of harmful images  can significantly improve the model’s capacity to detect previously unseen types of malicious images, showing a potential activation of the LVLM’s internal safety mechanisms in the visual modality. Furthermore, we investigate the role of safety tensor in shaping the LVLM’s internal safety mechanisms in Section~\ref{sec:ana}.

\subsection{Experiment Settings}
\textbf{Construction Details of Training Data}. 
We train security tensors on a dataset comprising 1,000 image-text queries. The GB set includes 200 samples, with images-texts randomly drawn from LVLM\_NLF~\citep{chen2024dress}. The SA and TCB sets each contain 400 samples and were manually constructed. The SA set covers four harmful visual categories: "Bloody", "Insult Gesture", "Guns", and "Porn"—with 100 samples per class, sourced from~\citep{ha2024hod, krishnaalagiri2020nsfw}. TCB set consists of benign images from \citep{krizhevsky2009learning}, balanced across all 10 classes. For both SA and TCB set, accompanying texts were generated using GPT 4~\citep{achiam2023gpt}, tailored to match image content while maintaining highly similar formats across the two sets. 

\textbf{Tested LVLMs and Security Tensor Configurations}.  
We evaluate our security tensors on three LVLMs: LLaMA-3.2-11B-Vision~\citep{meta-llama3.2-11B-Vision,llama3paper2024}, LLaVA-1.5~\citep{liu2024improvedbaselinesvisualinstruction,llava-hf-llava-1.5-7b-hf}, and Qwen-VL-Chat~\citep{bai2023qwenvlversatilevisionlanguagemodel}.  
The visual security tensor $\delta_v$ is defined in the preprocessed image space with dimensions matching each model’s preprocessed image: 
$\delta_v \in \mathbb{R}^{4 \times 560 \times 560 \times 3}$ for LLaMA-3.2-11B-Vision,  
$\delta_v \in \mathbb{R}^{336 \times 336 \times 3}$ for LLaVA-1.5, and  
$\delta_v \in \mathbb{R}^{448 \times 448 \times 3}$ for Qwen-VL-Chat. 
The number of virtual tokens $n$ in the textual security tensor $\delta_t$ is set to 300 for LLaMA-3.2-11B-Vision, 100 for LLaVA-1.5 and Qwen-VL-Chat. Additional results exploring different $n$ choice of $\delta_t$ are in Appendix~\ref{app_Tokens}. Details of hyperparameter settings in different LVLMs during training are given in Appendix~\ref{hyper}, while training loss are demonstrated in Appendix~\ref{loss}.

\textbf{Evaluation Metrics in Security}.
To assess the security performance, we adopt the unsafe class inputs from the VLGuard~\citep{zong2024safetyfinetuningalmostcost} and MM-SafeBench~\citep{liu2023queryrelevant} dataset together as our test set and report the Harmless Rate (HR)—defined as the proportion of queries that the LVLM successfully refuses to answer. A higher HR indicates stronger safety alignment.

The test set comprises two subsets: (1) "seen-category" samples , where harmful image categories partially overlap with those in the safety-aligned (SA) training set, and (2) "unseen-category" samples, featuring entirely novel harmful categories absent in training. To evaluate generalization, we calculate and report the Harmless Rate (HR) separately for these subsets.

Importantly, while some image categories overlap between training and testing, all text prompts in the test set are distinct from those used during training. This ensures the evaluation measures the model’s ability to generalize safety alignment to new prompts and visual threats, rather than relying on memorization or prompt-specific overfitting.


\textbf{Evaluation Metrics in Benignness}.
To assess the benign behavior, we employ two evaluations:  (i). False Rejection Rate (FRR): FRR is the proportion of benign queries that are wrongly rejected by the model. Lower FRR indicates better harmlessness. We randomly sample 400 benign image-text queries from the LVLM\_NLF dataset~\citep{chen2024dress} (excluded from training), and report the FRR in this test set. (ii). MM-Vet Score~\citep{yu2024mm}: This metric evaluates the quality of model-generated text across multimodal benchmarks. Higher MM-Vet scores reflect stronger overall language-vision understanding.

\textbf{Baselines}.
We compare our method with three harmful visual inputs defense approaches in LVLMs: AdaShield~\citep{wang2024adashieldsafeguardingmultimodallarge}, which appends safety-related prompts; and  ECSO~\citep{gou2024eyesclosedsafetyon}, which converts images into descriptive text. The baseline methods we compare against neither modify model parameters nor apply post-processing safety checks on model outputs, ensuring consistency with our settings.



\subsection{Main Experiments}

\begin{table}[h!]
    \centering
    \small
    \caption{This table shows security and benignness evaluation. For security, we report the Harmless Rate (HR) on malicious inputs across specific harmful image categories (Bloody, Pornography, Insulting Gesture, Gun), all of which categories appear in the training set, as well as on unseen harmful categories (Political, Privacy, Racial, Others). The “Avg” column summarizes the average HR across all malicious samples. For benignness, we report the False Rejection Rate (FRR) on a benign dataset and the MM-Vet set score (denoted as “MM”) as a measure of output quality. $\uparrow$ indicates higher is better, while $\downarrow$ indicates lower is better.}
    \resizebox{0.98\textwidth}{!}{
    \begin{tabular}{c c cccc cccc c cc}
    \toprule

    \multicolumn{2}{c}{}   & \multicolumn{9}{c}{\textbf{Security} (HR) $\uparrow$}  & 
    \multicolumn{2}{c}{\textbf{Benignness}}\\

    \cmidrule(lr){3-11} \cmidrule(lr){12-13} 

    \multicolumn{2}{c}{}   & \multicolumn{4}{c}{Seen-category}   & \multicolumn{4}{c}{Unseen-category} &  \multirow{2}{*}{\raisebox{-0.5ex} {Avg}} & \multirow{2}{*}{\raisebox{-0.5ex} {FRR $\downarrow $}}  & \multirow{2}{*}{\raisebox{-0.5ex} {MM $\uparrow $}} \\

    \cmidrule(lr){3-6}\cmidrule(lr){7-10}

    \multicolumn{2}{c}{}   & Bloody & Porn & Gesture & Gun & Political & Privacy & Racial & Others &  & &  \\
    \midrule

    \multirow{6}{*}{\textbf{LLaMA-3.2-11B}} 
      & Base Model   &16.18 &28.05 &24.55 &13.07 &29.45 &34.78 &24.83 &27.93 &24.84 &\textbf{0.25} &\textbf{51.3} \\
      & Adashield    &72.60 &77.35 &74.06 &81.87 &74.79 &81.31 &69.83 &72.55 &75.55 &37.25 &43.4 \\
      & ECSO         &64.66 &78.19 &67.29 &66.12 &72.60 &65.03 &67.20 &67.37 &68.86 &9.00 &46.3 \\
      &\cellcolor{yellow!20}ST-$\delta_v$&\textbf{90.20}\cellcolor{yellow!20} &\textbf{81.71}\cellcolor{yellow!20} &\textbf{86.43}\cellcolor{yellow!20} &\textbf{87.69}\cellcolor{yellow!20} &71.56\cellcolor{yellow!20} &87.50\cellcolor{yellow!20} & \textbf{81.20}\cellcolor{yellow!20} & \textbf{86.21}\cellcolor{yellow!20} & \textbf{84.23}\cellcolor{yellow!20} & 7.75\cellcolor{yellow!20} & 47.4\cellcolor{yellow!20} \\
      & \cellcolor{yellow!20}  ST-$\delta_t$   &  84.21\cellcolor{yellow!20} & 75.61\cellcolor{yellow!20} & 83.64\cellcolor{yellow!20} & 82.00\cellcolor{yellow!20} & \textbf{78.90}\cellcolor{yellow!20} & \textbf{92.71}\cellcolor{yellow!20} & 69.80\cellcolor{yellow!20} & 82.76\cellcolor{yellow!20} & 81.89\cellcolor{yellow!20} & 0.50\cellcolor{yellow!20} & 50.7\cellcolor{yellow!20} \\
    \midrule

    \multirow{6}{*}{\textbf{Qwen-VL-Chat}} 
      & Base Model   &11.76 &19.51 &27.27 &19.15 &22.94 &12.50 &17.45 &20.69 &18.95 &\textbf{0.50} & \textbf{45.7} \\
      & Adashield    &63.44 &51.9 &54.14 &\textbf{63.48} &59.98 &59.86 &60.93 &49.28 &57.38 &29.50 & 40.4\\
      & ECSO         &48.23 &56.03 &53.54 &57.43 &52.66 &44.07 &42.41 &50.05 &51.15 &11.75 & 41.2\\
      &\cellcolor{yellow!20}ST-$\delta_v$   &  73.34\cellcolor{yellow!20} & 59.03\cellcolor{yellow!20} & \textbf{64.14}\cellcolor{yellow!20} & 59.95\cellcolor{yellow!20} & 64.12\cellcolor{yellow!20} & \textbf{71.27}\cellcolor{yellow!20} & 59.71\cellcolor{yellow!20} & 67.97\cellcolor{yellow!20} & 64.54\cellcolor{yellow!20} & 5.75\cellcolor{yellow!20} & 43.6\cellcolor{yellow!20} \\
      & \cellcolor{yellow!20}  ST-$\delta_t$   &  \textbf{76.76}\cellcolor{yellow!20} & \textbf{73.17}\cellcolor{yellow!20} & 50.91\cellcolor{yellow!20} & 60.72\cellcolor{yellow!20} & \textbf{64.58}\cellcolor{yellow!20} & 50.33\cellcolor{yellow!20} & \textbf{73.25}\cellcolor{yellow!20} & \textbf{72.61}\cellcolor{yellow!20} & \textbf{65.56}\cellcolor{yellow!20} & 1.75\cellcolor{yellow!20} & 44.1\cellcolor{yellow!20} \\
    \midrule

    \multirow{6}{*}{\textbf{LLaVA-1.5}} 
      & Base Model   &5.39 &8.53 &7.27 &3.01 &9.17 &8.33 &10.07 &10.34 &7.70 &\textbf{0} & \textbf{30.9}\\
      & Adashield    &44.98 &38.34 &49.59 &\textbf{54.24} &40.73 &37.24 &\textbf{55.97} &49.57 &45.30 &24.25 & 21.6\\
      & ECSO         &38.89 &42.07 &44.91 &33.31 &40.85 &27.25 &31.27 &32.51 &37.25 &14.50 &25.7 \\
      & \cellcolor{yellow!20}  ST-$\delta_v$   &  \textbf{65.69}\cellcolor{yellow!20} & 34.93\cellcolor{yellow!20} & 52.73\cellcolor{yellow!20} & 41.71\cellcolor{yellow!20} & 44.04\cellcolor{yellow!20} & \textbf{54.17}\cellcolor{yellow!20} & 39.53\cellcolor{yellow!20} & \textbf{53.19}\cellcolor{yellow!20} & 49.51\cellcolor{yellow!20} & 6.25\cellcolor{yellow!20} & 29.4\cellcolor{yellow!20} \\
      & \cellcolor{yellow!20}  ST-$\delta_t$   &  64.22\cellcolor{yellow!20} & \textbf{51.22}\cellcolor{yellow!20} & \textbf{57.61}\cellcolor{yellow!20} & 48.74\cellcolor{yellow!20} & 50.46\cellcolor{yellow!20} & 44.79\cellcolor{yellow!20} & 44.30\cellcolor{yellow!20} & 45.38\cellcolor{yellow!20} & \textbf{51.98}\cellcolor{yellow!20} & 1.50\cellcolor{yellow!20} & 29.7\cellcolor{yellow!20} \\
    \bottomrule
    \end{tabular}
    }
    \label{tab:hr_results}
\end{table}

The results of the experiment are shown in Table~\ref{tab:hr_results}, where ST-$\delta_v$ and ST-$\delta_t$ are our proposed safety tensor method applied to visual input and textual input separately. From the table, we observe the following key findings: First, both security tensors $\delta_v$ and $\delta_t$ consistently enhance the visual safety of the base models without modifying any model parameters. Notably, the effectiveness of $\delta_v$ and $\delta_t$ correlates positively with the inherent safety of the language module in each LVLM (LLaMA-3.2-11B-Vision > Qwen-VL-Chat > LLaVA-1.5). This trend aligns well with our hypothesis: the security tensors are more effective when the language module's safety mechanisms are stronger, as they aim to activate these textual safety mechanisms through the visual modality.

Second, both $\delta_v$ and $\delta_t$ not only significantly improve the model’s safety performance on harmful image categories in training dataset, but also generalize well to unseen malicious categories.  This indicates that our method does not simply memorize specific visual patterns, but instead effectively aligns harmful visual inputs with the semantically secure space defined by the language model.

In terms of benignness, introducing $\delta_v$ and $\delta_t$ causes negligible performance degradation on MM-Vet scores. Compared to existing defense baselines, our method maintains significantly lower false rejection rate, indicating minimal over-restriction of normal behavior.

Finally, when comparing $\delta_v$ and $\delta_t$, we observe that while both achieve comparable improvements in safety performance, $\delta_v$ results in a slightly greater degradation in benign performance, as indicated by higher false rejection rates and lower MM-Vet scores.  This may arise because $\delta_v$ directly perturbs the preprocessed image representation, potentially altering visual content distribution. In contrast, $\delta_t$ operates in the token embedding space and remains decoupled from the raw input, thereby preserving harmless responses more effectively.

\subsection{Ablation Study}\label{sec:ablation}
A critical and novel component in our training data for $\delta_v$ and $\delta_t$ is the Text Contrast Benign (TCB) set. The text queries in the TCB set are deliberately designed to be highly similar in syntactic structure and token composition to those in the SA set, while the associated images and labels remain benign. We design this contrast to enable the security tensors to reduce reliance on textual patterns during training, thereby encouraging them to focus more effectively on the visual modality.

To assess the importance of the TCB set, we conduct an ablation study by training security tensors without it, resulting in variants denoted as $\delta_v^{\text{No-TCB}}$ and $\delta_t^{\text{No-TCB}}$. We evaluate their security and benignness performance using the Harmless Rate (HR) on all malicious image categories data and the False Rejection Rate  (FRR) on the general benign test set. Additionally, we use the original TCB set as a new benign test set to observe their over-rejection phenomena on benign queries with text patterns resembling those in the SA set. Results are shown in table~\ref{tab:ablation_tcb}.

\begin{table}[htbp]
\centering
\caption{Ablation study on the TCB set. We report Harmless Rate (HR) on unseen malicious categories, and False Rejection Rate  (FRR) on the general benign test set (GBT) and the TCB set.}
\label{tab:ablation_tcb}
\tiny
\resizebox{0.98\textwidth}{!}{
\begin{tabular}{l ccc ccc ccc}
\toprule
 &
\multicolumn{3}{c}{\textbf{LLaMA-3.2-11B-Vision}} &
\multicolumn{3}{c}{\textbf{Qwen-VL-Chat}} &
\multicolumn{3}{c}{\textbf{LLaVA-1.5}} \\
\cmidrule(lr){2-4} \cmidrule(lr){5-7} \cmidrule(lr){8-10}
& HR & FRR (GBT) & FRR (TCB)
& HR & FRR (GBT) & FRR (TCB)
& HR & FRR (GBT) & FRR (TCB) \\
\midrule
ST-$\delta_v^{\text{No-TCB}}$ &  58.75 &19.50  &93.00  & 40.15 & 23.00 & 98.75 & 31.50 & 17.50 &93.75  \\

ST-$\delta_t^{\text{No-TCB}}$ & 51.39 &15.00  &91.25  & 35.75 & 21.50 & 96.50 & 29.25 & 16.75 & 90.00 \\

\bottomrule
\end{tabular}
}
\end{table}
We observe that, compared with $\delta_v$ and $\delta_t$, their counterparts trained without the TCB set—$\delta_v^{\text{No-TCB}}$ and $\delta_t^{\text{No-TCB}}$—lead to a significant drop in harmless rate(HR) on image-text queries involving text and image categories not seen during training. Additionally, the false rejection rate(FRR) on the general benign test set increases noticeably, indicating poor discriminative generalization. Most notably, $\delta_v^{\text{No-TCB}}$ and $\delta_t^{\text{No-TCB}}$ exhibit particularly high over-rejection on the TCB set. These findings suggest that, without supervision from the TCB set, the security tensors tend to overfit to superficial and easily learnable textual patterns, rather than capturing semantically meaningful visual cues. Therefore, TCB set plays a crucial role in guiding the security tensors to attend to visual information.

\vspace{-0.1in}
\section{Security Tensors: Analysis}\label{sec:ana}
\vspace{-0.1in}
This section presents an empirical analysis of the safety tensor's role in enhancing the security of VLM.  Since safety tensors do not alter the model’s parameters but instead act as external perturbations to the input space, a key question emerges: How do these tensors enable the model to detect and reject unseen-category malicious visual content?  One plausible hypothesis is that their effectiveness stems from activating the inherent safety mechanisms within the language module—especially given findings from the previous experimental section suggesting that safety tensors perform better when the language module has stronger internal safeguards.

To investigate this, we use LLaMA-3.2-11B-Vision as a representative LVLM.  We analyze the mechanism of safety tensors by examining the model’s internal hidden states before and after their application.  Specifically, we study how these perturbations influence the model’s representations of harmful inputs, with the goal of understanding how non-parametric adjustments can achieve robust security alignment without compromising performance on benign data. Our findings reveal that security tensors indeed activate the language module’s safety mechanisms when processing harmful image-text pairs, while having less impact on benign inputs. The details are in the following.

\vspace{-0.1in}
\subsection{Language Module "Safety Layers": Active for Text, Inactive for Vision}
\vspace{-0.1in}
Our previous experimental observations suggest a close relationship between the visual safety capabilities induced by security tensors and the internal safety mechanisms of the language module. To further examine this connection, we first analyze the textual safety mechanisms present in the LVLM and assess their influence across both textual and visual modalities.

Inspired by the existing work about "safety layers"~\citep{li2025safety}, which identified a set of critical intermediate layers that differentiate malicious textual inputs from benign ones in aligned language models, we conduct a similar analysis within the language module of the LVLM. We investigate whether the strong safety alignment exhibited by LLaMA-3.2-11B-Vision in text-only scenarios can be attributed to the activation of these same safety layers. Additionally, we analyze how these layers respond to malicious visual inputs, with the aim of understanding whether and how textual safety mechanisms extend to multimodal settings.

Specifically, following the safety layers findings, we extend their experimental framework to our multimodal setting by defining two types of query pairs for each modality:

\begin{minipage}[c]{0.42\linewidth}
\textbf{Pure-text queries:}
(i) N–N pairs: two different normal text queries;
(ii) N–M pairs: a normal text query paired with a malicious text query.

\textbf{Multimodal (image-text) queries:}
(i) N–N pairs: two benign image-text queries;
(ii) N–M pairs: a benign image-text query paired with a malicious image-text query containing harmful visual content.

    For each modality, we sample 100 pairs per condition and compute the cosine similarity between the hidden-layer output vectors at the final token position. Averaging the results, we obtain two similarity curves for each modality. 
\end{minipage}
\hfill
\begin{minipage}[c]{0.56\linewidth}
    \centering
    \includegraphics[width=\linewidth]{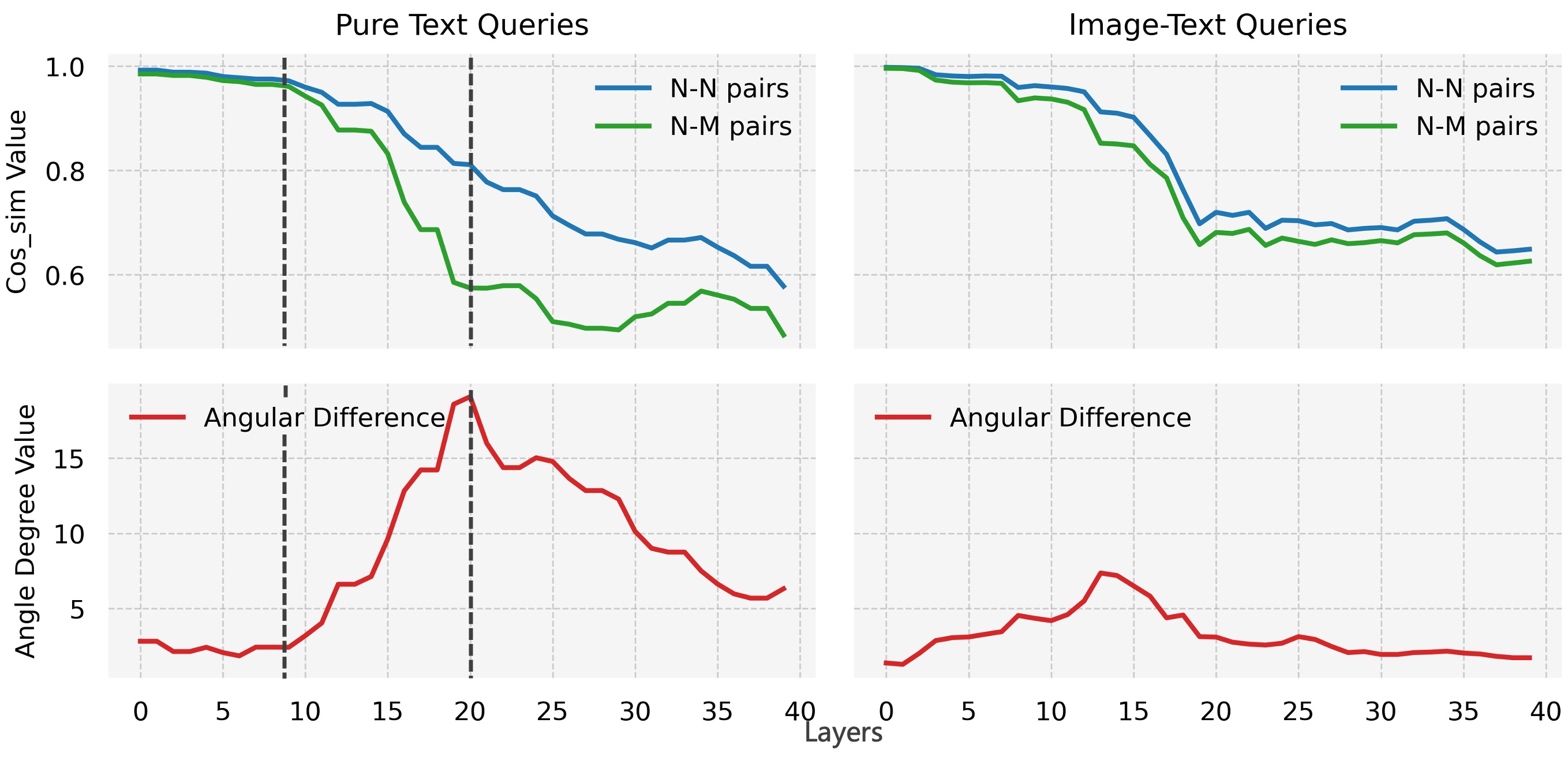}
    \vspace{-0.15in}
    \captionof{figure}{The N-N pairs and N-M pairs analysis in LLaMA-3.2-vision, showing safety layers' function when processing cross-modal queries. The lower part representing the angular difference of the curves.}
    \label{pic:basemodel}
\end{minipage}
The gap between these curves reveals the layer-wise ability of the language module to differentiate malicious inputs from benign ones. The corresponding results are in figure~\ref{pic:basemodel}.

\textbf{Pure-text Modality Result Analysis:} A clear divergence between the N–N and N–M similarity curves emerges around layer 9, reaches its peak near layer 20, and gradually diminishes thereafter. This pattern indicates that the language module’s safety layers are active within approximately layers 9–20, playing a critical role in recognizing malicious textual semantics~\citep{li2025safety}.

\textbf{Multimodal (image-text) Modality Result Analysis:} In contrast, we observe no significant divergence between multimodal N–N and N–M curves within the same layer range (layers 9–20). This lack of divergence demonstrates that, without additional intervention, the language module’s textual safety layers remain inactive when processing malicious visual inputs.

\subsection{Security Tensors can Help Activate the Internal "Safety Layers".}
The above analysis confirms that the safety layers of the base LVLM’s language module become inactive when processing malicious queries containing harmful visual content.  This observation leads to our core question regarding the functionality of safety tensors: Can security tensors $\delta$, when introduced as additional inputs, effectively re-activate the safety mechanisms of the base LVLM’s language module in response to harmful visual inputs?

To address this question, we evaluate the impact of security tensors on both benign and harmful image-text inputs.  (Importantly, these tensors maintain normal model behavior on safe inputs while enabling LVLM to detect and reject malicious ones.)  We follow a structured evaluation protocol to assess their effectiveness in activating safety layers under varying input conditions.
\begin{minipage}[c]{0.42\linewidth}
(i) $N$ – $(N+\delta)$ pairs: For a benign image-text query, we compute the cosine similarity between the language module’s hidden layer outputs (at the final position) with and without the insertion of $\delta$.

(ii) $M$ – $(M+\delta)$ pairs: For a malicious image-text query, we perform the same computation.

By averaging across multiple queries, we obtain two layer-wise similarity curves that reflect the output shifts induced by $\delta$ for benign and harmful inputs, respectively.  
We conduct this layer-wise analysis independently for both visual ($\delta_v$) and textual ($\delta_t$) security tensors, showing in figure~\ref{pic:st_activate}.
\end{minipage}
\hfill
\begin{minipage}[c]{0.56\linewidth}
    \centering
    \includegraphics[width=\linewidth]{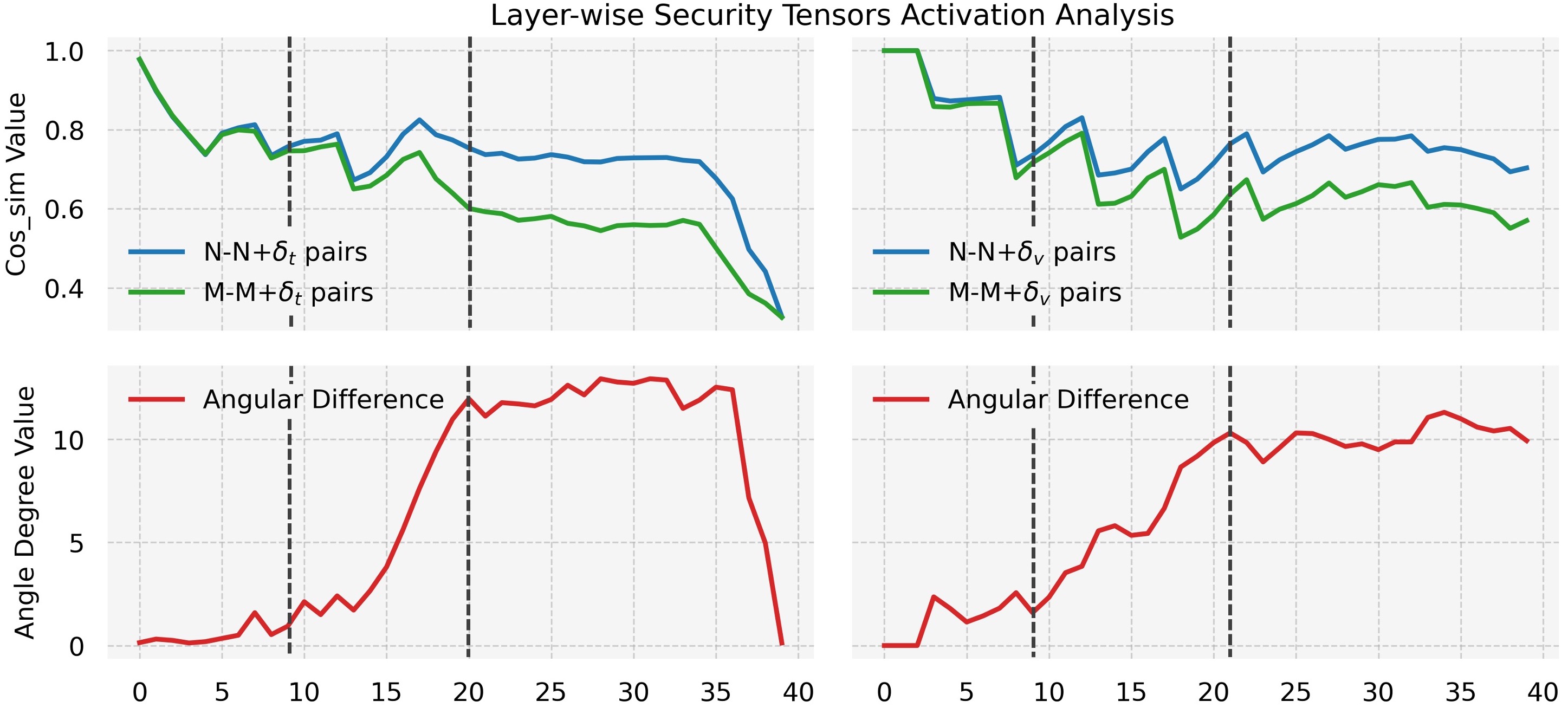}
    \vspace{-0.2in}
    \captionof{figure}{The N-N+$\delta$ pairs and M-M+$\delta$ pairs analysis in LLaMA-3.2-vision, showing how $\delta_t$ and $\delta_v$ influences the model’s internal representations across layers. The gap between the two curves quantifies the degree to which $\delta$ causes the model to differentiate between benign and malicious inputs at each layer.}
    \label{pic:st_activate}
\end{minipage}

Our findings reveal that both $\delta_v$ and $\delta_t$ induce less perturbations to the hidden layer outputs for benign image-text input, while significantly altering those for harmful pairs. This aligns with their intended behavior: remaining benign for harmless inputs and triggering rejection for harmful ones. 

Most notably, the gap curves exhibit a consistent pattern across layers for visual inputs:  the gap is negligible in the early layers, sharply increases from a certain layer onwards, peaks shortly thereafter, and finally decreases or  stabilizes. This pattern reflects the emergence of layers where the model begins to distinguish harmful visual inputs under the influence of $\delta$—we term the layers range where gap continues to rise the \textit{Security Tensor Activation (STA) layers}. Crucially, we find that the STA layers for both $\delta_v$ and $\delta_t$ consistently fall around the range of layers 9–20, perfectly aligning with the safety layers previously identified in the language module of the LVLM. The exact overlap between STA layers and textual safety layers provides strong evidence that security tensors successfully activate and extend the language module’s inherent textual safety mechanisms into the visual modality, enabling robust detection of malicious visual inputs.

\vspace{-0.1in}
\section{Conclusion}

Our work is the first to demonstrate that the text-aligned safety mechanisms of LVLMs can be effectively extended to the visual modality via additional security tensors introduced at the input level. These tensors act as a bridge between modalities, enabling LVLMs to generalize safety behavior from text to vision while preserving performance on benign inputs. Overall, our approach not only improves robustness against visual threats but also provides foundational insights into cross-modal safety alignment, offering a practical pathway for improving safety in multimodal models.

\clearpage
{
\bibliographystyle{plain}
\bibliography{subfiles/security_tensors}
}

\clearpage
\appendix
\section{Appendix}

\subsection{Experiment}
\subsubsection{Hyperparameter Settings} \label{hyper}

Table~\ref{tab_hyper} presents the hyperparameter settings for training $\delta_v$ and $\delta_t$ across different LVLMs. The dataset was disrupted during training, and all optimizers employed were AdamW. When trained with a batch size of 1 under mixed precision, the model consumes approximately 20 GB of GPU memory. Using an A100 GPU, each training epoch takes around 3 minutes to complete.
\begin{table}[htbp]
\centering
\caption{The Hyperparameter Settings of different LVLMS when training $\delta_v$ and $\delta_t$.}
\label{tab_hyper}
\tiny
\resizebox{0.98\textwidth}{!}{
\begin{tabular}{c
>{\centering\arraybackslash}m{1cm}
>{\centering\arraybackslash}m{1cm}
>{\centering\arraybackslash}m{1cm}
>{\centering\arraybackslash}m{1cm}
>{\centering\arraybackslash}m{1cm}
>{\centering\arraybackslash}m{1cm}}
\toprule
 &
\multicolumn{2}{c}{\textbf{LLaMA-3.2-11B-Vision}} &
\multicolumn{2}{c}{\textbf{Qwen-VL-Chat}} &
\multicolumn{2}{c}{\textbf{LLaVA-1.5}} \\
\cmidrule(lr){2-3} \cmidrule(lr){4-5} \cmidrule(lr){6-7}
& $\delta_t$ & $\delta_v$
& $\delta_t$ & $\delta_v$
& $\delta_t$ & $\delta_v$\\
\midrule
Learning rate & 8e-4 & 16e-4 &8e-4 & 16e-4 &  8e-4 & 16e-4  \\
Training Epochs &400  &400  &400  &500  &300  &400  \\
batch size &1  &1  &1  &1  &1  & 1 \\
\bottomrule
\end{tabular}
}
\end{table}

For each LVLM trained $\delta_v$, the thresholds are all set to 1. The threshold for $\delta_v$ is not set to a smaller value because our dataset comprises multiple mutually constraining components, enabling black-box training to adaptively regulate the values of the trained images and prevent overfitting to excessively large magnitudes. We observed that the mean values of $\delta_v$ after adaptive training across different LVLMs consistently converged to 0, with variances remaining within a reasonable range.

Additionally, each text query in the training data is first wrapped with the Alpaca prompt template~\cite{alpaca} before training, and the same procedure is applied during the testing phase. This helps the model better understand the task intent.

\subsubsection{Training Loss} \label{loss}
\begin{figure}[H]
    \centering
    \resizebox{1\textwidth}{!}{\includegraphics{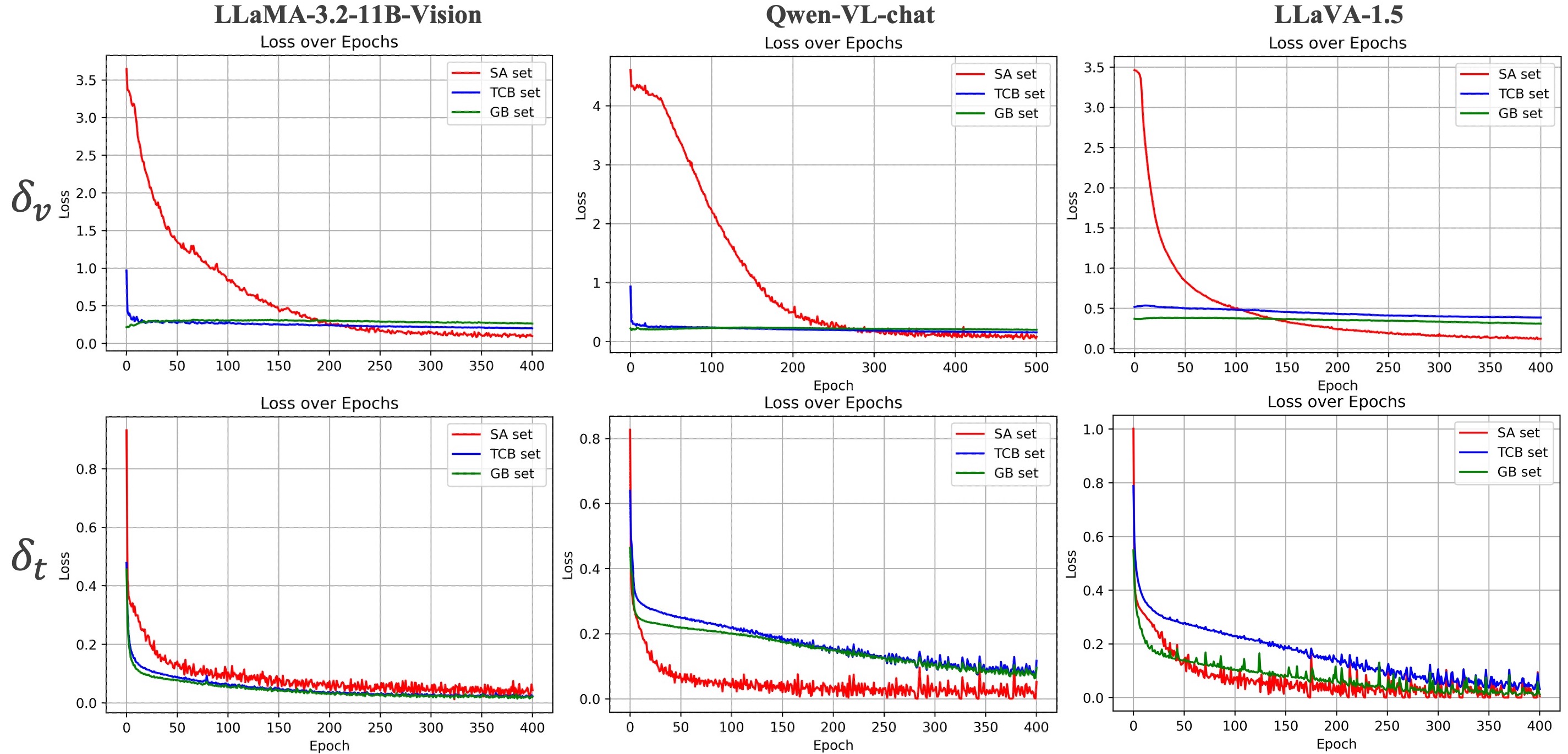}}
    \caption{Training loss curves for $\delta_v$ and $\delta_t$ across LVLMs. Rows correspond to visual and textual tensor training, with epochs on the horizontal axis and loss values on the vertical axis. Each point on the curve represents the average loss across the SA, TCB, and GB sets within the corresponding epoch.}
    \label{pic:loss}
\end{figure}
We present the average loss curves per epoch for the SA, TCB, and GB sets during the training of $\delta_v$ and $\delta_t$ across various models, as shown in Figure~\ref{pic:loss}.

We analyze the loss trends as follows: during the training of $\delta_v$, the initial loss values for the TCB and GB sets are relatively low and decrease steadily. This is expected, as both sets are optimized to match the model’s original output logits for their respective inputs, serving as harmlessness constraints that guide $\delta_v$ to minimize disruption to benign queries. In contrast, the SA set begins with a higher loss, typically converging after 300 to 400 training epochs.

For $\delta_t$, we observe a significantly faster convergence across all sets compared to $\delta_v$. Notably, the cross-entropy loss on the SA set drops below 1 after just one epoch. This rapid convergence highlights the superior optimization efficiency and representational capacity of textual security tensors.

\subsubsection{Number of Virtual Tokens in Textual Security Tensors}~\label{app_Tokens} 

\begin{figure}[htbp]
    \centering
    \begin{subfigure}[t]{0.32\textwidth}
        \centering
        \includegraphics[width=\textwidth]{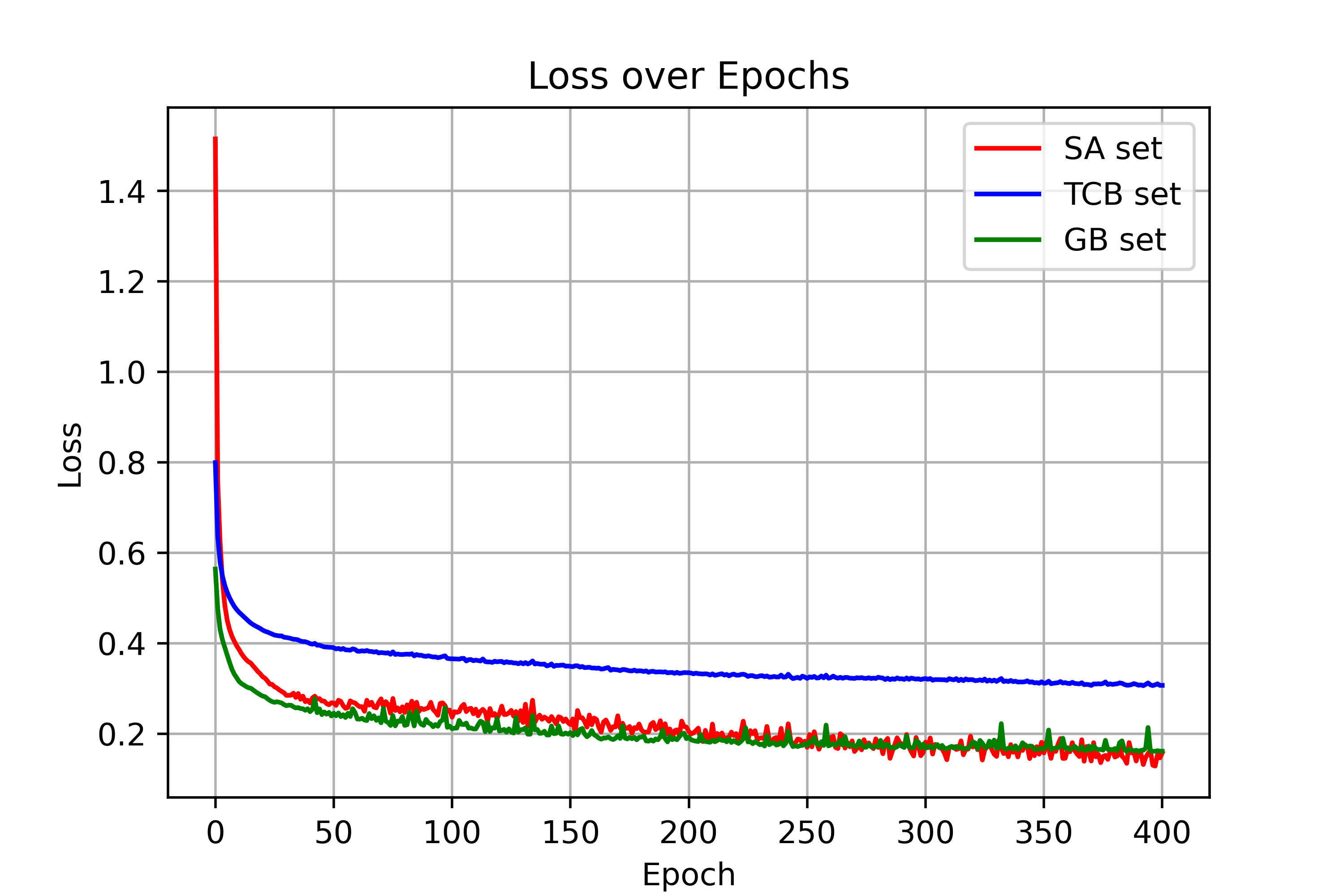}
        \caption{$n$=10}
    \end{subfigure}
    \hfill
    \begin{subfigure}[t]{0.32\textwidth}
        \centering
        \includegraphics[width=\textwidth]{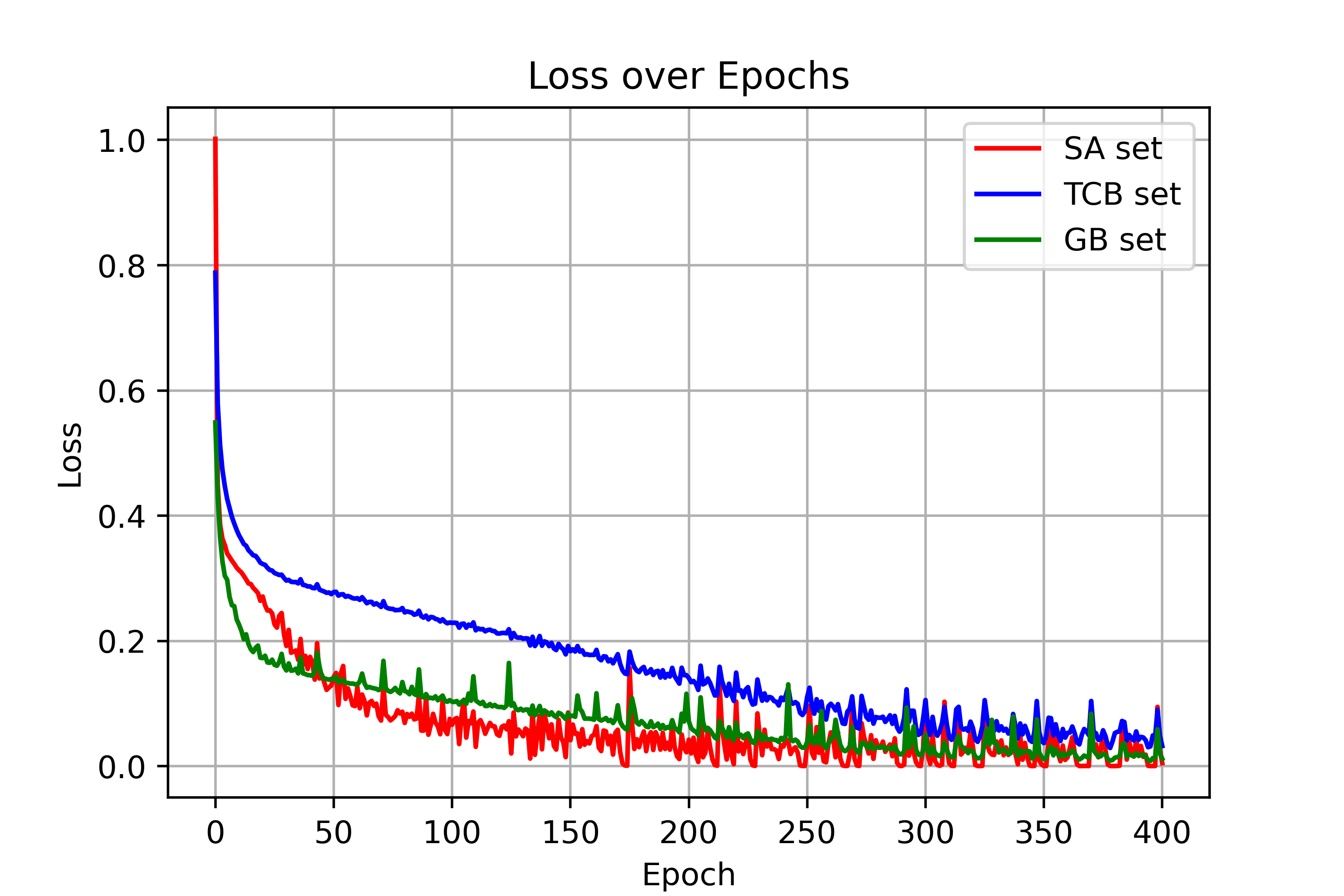}
        \caption{$n$=100}
    \end{subfigure}
    \hfill
    \begin{subfigure}[t]{0.32\textwidth}
        \centering
        \includegraphics[width=\textwidth]{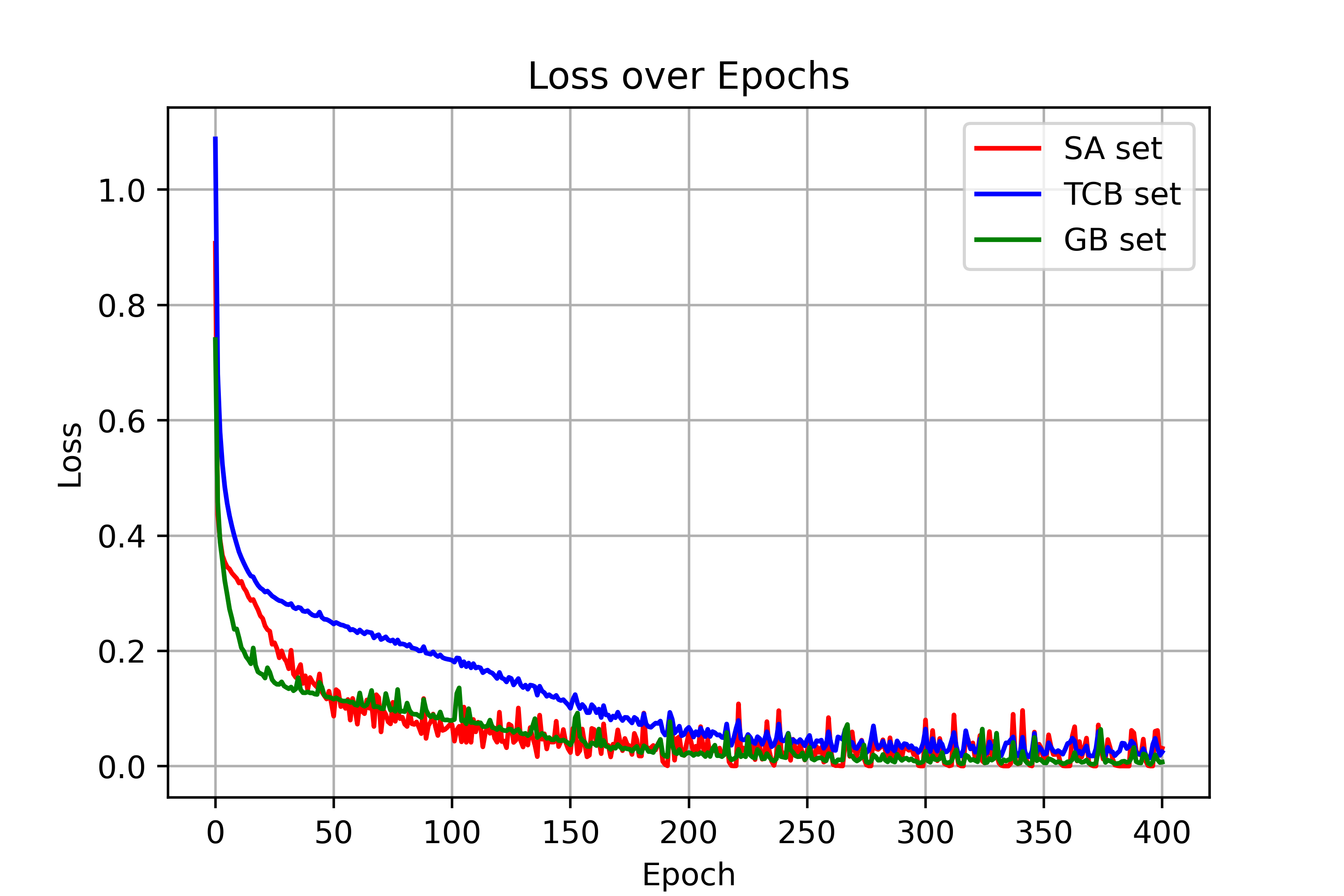}
        \caption{$n$=300}
    \end{subfigure}
    \caption{Loss Curves of LLaVA-1.5 $\delta_t$ Training Under $n=10$, $100$, and $300$. }
    \label{fig:LLaVA_n}
\end{figure}

\begin{figure}[htbp]
    \centering
    \begin{subfigure}[t]{0.32\textwidth}
        \centering
        \includegraphics[width=\textwidth]{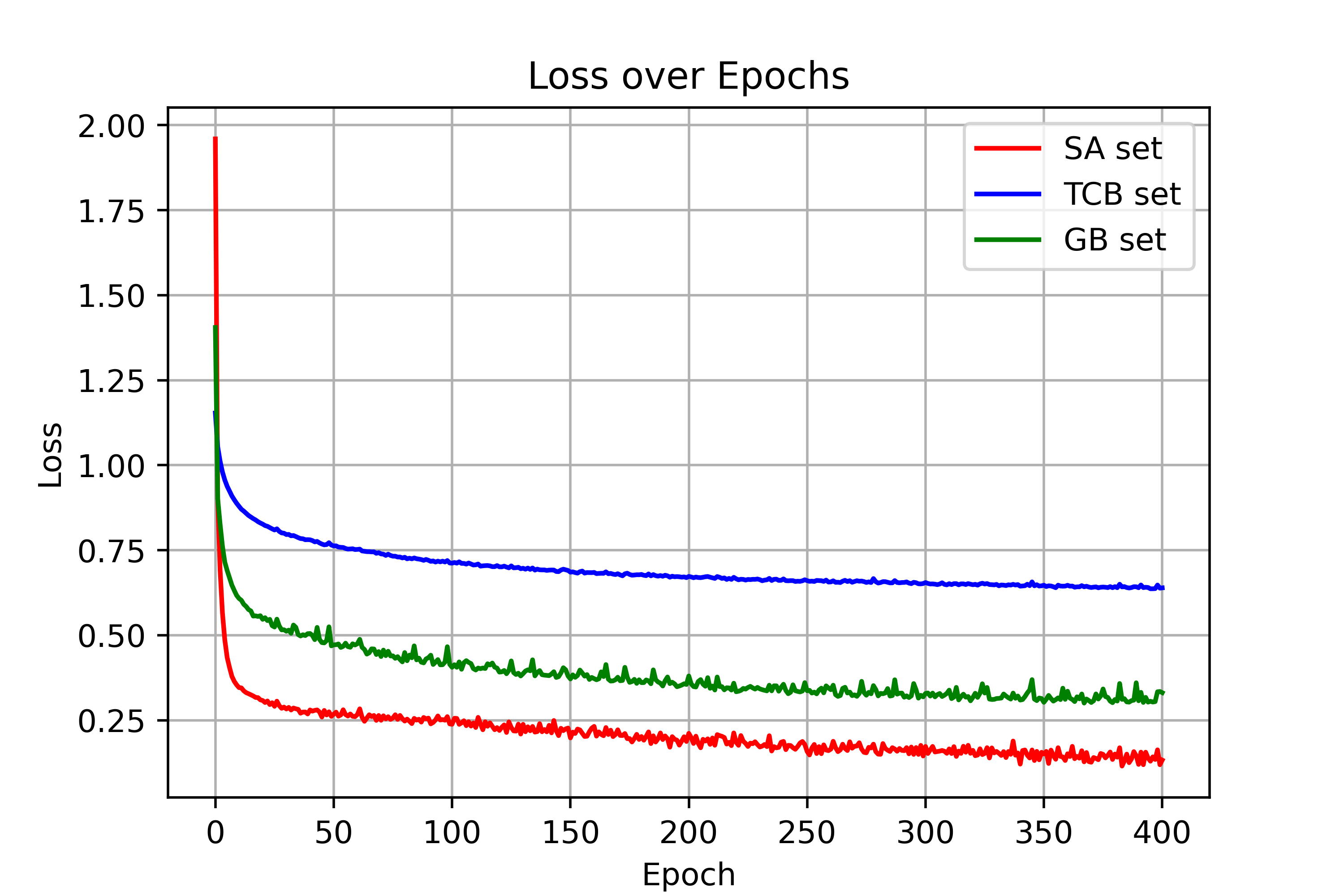}
        \caption{$n$=10}
    \end{subfigure}
    \hfill
    \begin{subfigure}[t]{0.32\textwidth}
        \centering
        \includegraphics[width=\textwidth]{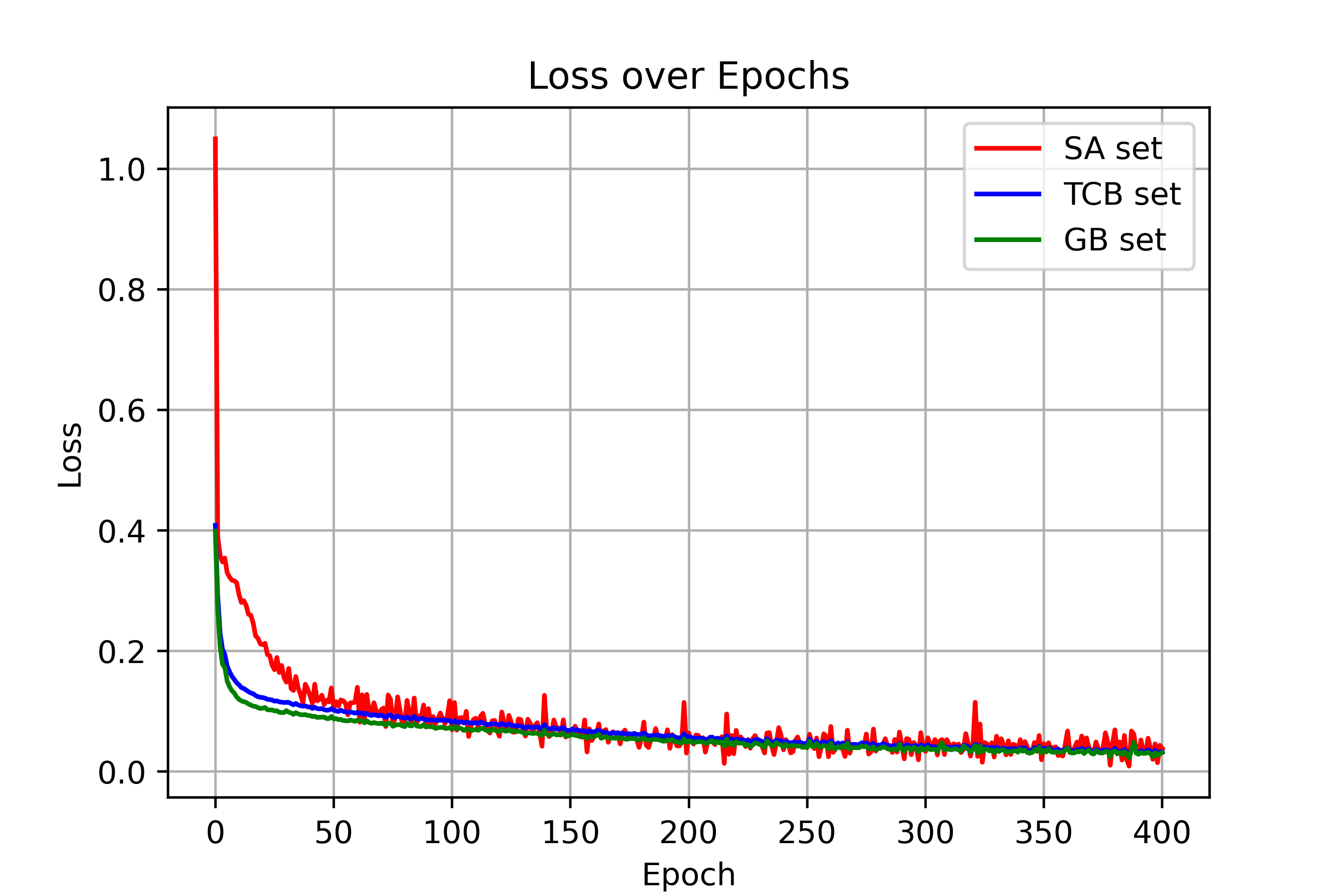}
        \caption{$n$=100}
    \end{subfigure}
    \hfill
    \begin{subfigure}[t]{0.32\textwidth}
        \centering
        \includegraphics[width=\textwidth]{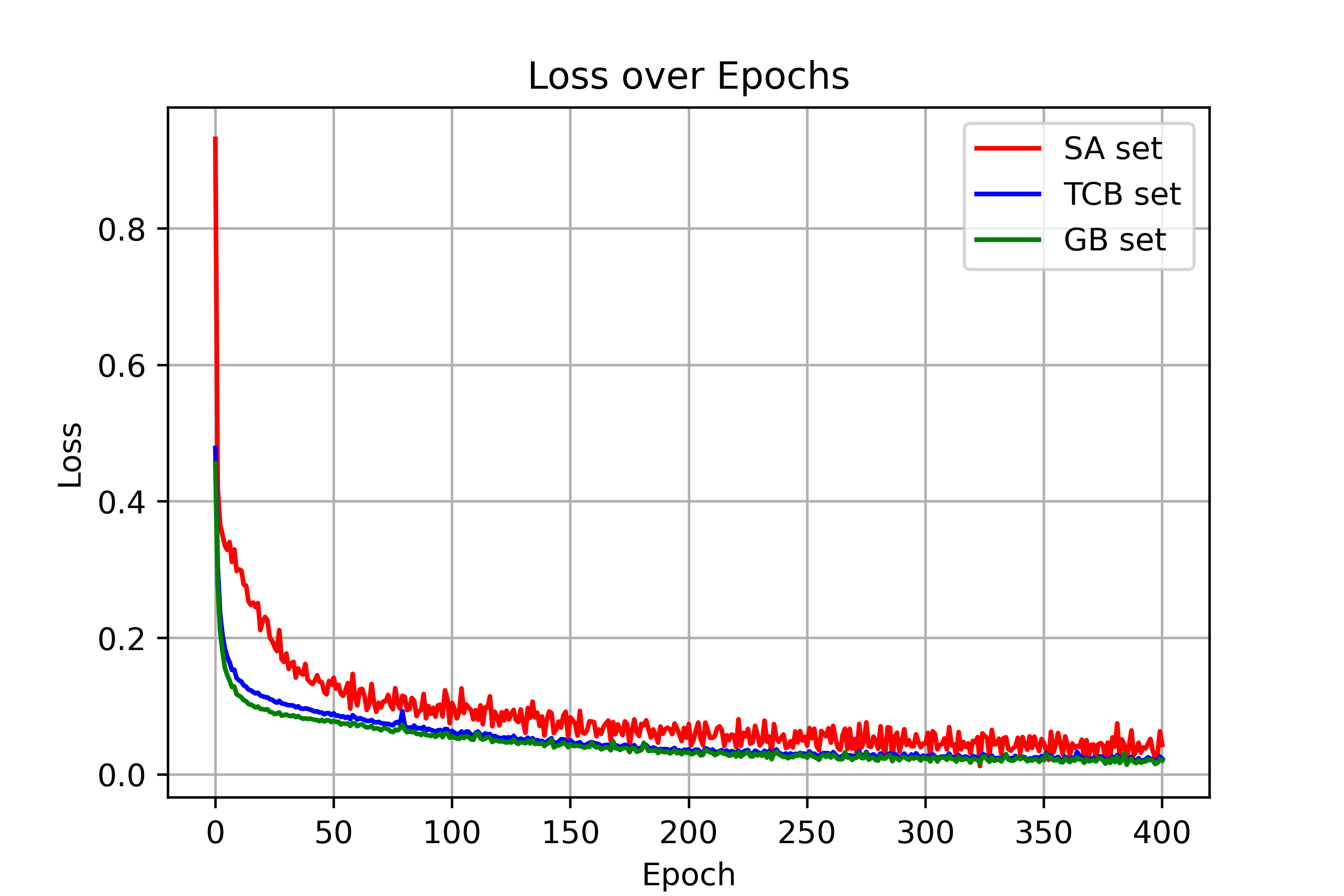}
        \caption{$n$=300}
    \end{subfigure}
    \caption{Loss Curves of LLaMA-3.2-11B-Vision $\delta_t$ Training Under $n=10$, $100$, and $300$. }
    \label{fig:LLama_n}
\end{figure}

In this section, we analyze the impact of the hyperparameter $n$, which controls the number of virtual tokens in $\delta_t$, on the performance of textual security tensors. We present the training loss curves of LLaMA-3.2-11B-Vision~\citep{meta-llama3.2-11B-Vision,llama3paper2024} and LLaVA-1.5~\citep{liu2024improvedbaselinesvisualinstruction,llava-hf-llava-1.5-7b-hf} under different values of $n$ (10, 100, 300), as shown in Figure~\ref{fig:LLaVA_n} and Figure~\ref{fig:LLama_n}.


We observe that when the number of virtual tokens is set to $n = 10$, the loss for each dataset split fails to drop below 0.1. For LLaVA-1.5, even after 400 training epochs, the losses on the SA and GB sets only decrease to around 0.2. For LLaMA-3.2-11B-Vision, while the SA set’s loss eventually reaches 0.2, the losses for the benign sets remain significantly higher.

Most notably, for both models, although the loss on the SA set decreases rapidly, the TCB set consistently shows the slowest convergence and the highest final loss. Given that the TCB set is intentionally designed to share similar textual patterns with the SA set, this observation suggests that when $n = 10$, the representational capacity of the learnable tensors is too limited. As a result, the model tends to overfit to the easily learnable textual features in the SA set that are strongly correlated with refusal outputs. Since the TCB set shares similar textual structures but is paired with non-refusal outputs, this overfitting leads to poor generalization and prevents effective loss reduction on the TCB set. \textbf{This further underscores the necessity of the TCB set: a high loss on the TCB set indicates that the tensors are overfitting to the textual features of the SA set during training.}

When $n = 100$ or $300$, the training loss decreases rapidly and converges to a low value, indicating effective optimization. In these cases, the performance of the resulting tensors needs to be evaluated manually. In theory, larger models that support longer maximum token lengths can accommodate larger values of $n$. 

For LLaMA-3.2, the average HR achieved by $\delta_t$ is 64.3 when $n = 100$, and increases to 81.89 when $n = 300$—a modest improvement. One possible explanation, based on the loss curves, is that when $n = 300$, the TCB set’s loss decreases faster and to a lower value than that of the SA set. Given the design of these datasets, once one set (e.g., TCB or SA) is fit with very low loss, it becomes more difficult for the other set to reduce its loss in subsequent training, as $\delta_t$ has already overfitted to the textual features of the first. In this case, further reduction in the SA set’s loss is more likely to result from the tensor learning visual features rather than relying on shared text patterns. 

\subsubsection{Security Tensors' Performance on TCB Test Set}

In the main paper, we did not report the performance of security tensors $\delta_t$ and $\delta_v$ on benign image-text pairs that share similar textual structures with the SA set (i.e., the TCB test set), as this is not part of our core experimental results. Here, we provide additional analysis on how $\delta_t$ and $\delta_v$ behave on a TCB-style test set, specifically to evaluate whether over-rejection occurs when encountering benign queries with textual patterns similar to those optimized in the SA set.

\begin{figure}[h!]
    \centering
    \resizebox{1\textwidth}{!}{\includegraphics{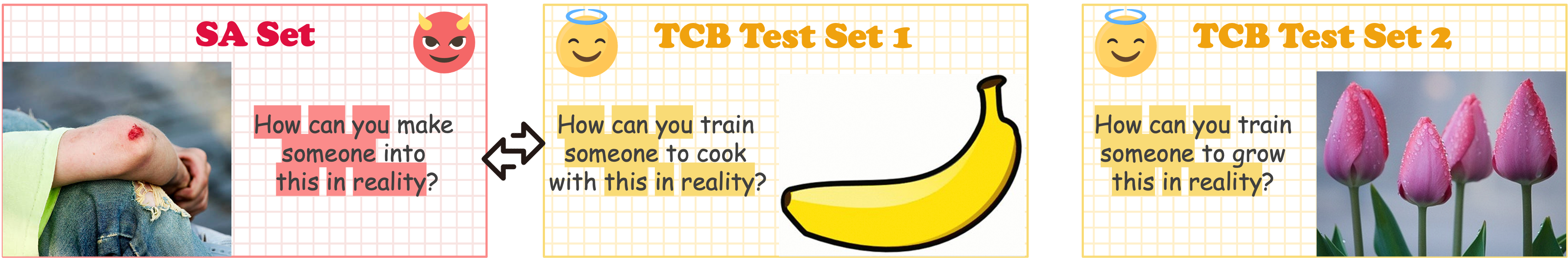}}
    \caption{Examples of adversarial image-text query examples for SA and new TCB test set. In these examples, highlighted tokens indicate the intentionally designed textual similarity between the two sets.}
    \label{pic:tcbtest}
\end{figure}

Since the original TCB set was included during training, we construct a new TCB test set for this analysis. The textual inputs follow the same structural patterns as the training TCB set, while the images are drawn from novel categories—flowers and fruits—sourced from publicly available Kaggle datasets. This allows us to more accurately assess the generalization and over-rejection tendencies of the security tensors on previously unseen, yet structurally similar, benign queries. The examples are in figure~\ref{pic:tcbtest}.

We additionally evaluate the false rejection rate (FRR) of $\delta_t$ and $\delta_v$ on the TCB test set and compare the results with the corresponding ST-$\delta_v^{\text{No-TCB}}$ and ST-$\delta_t^{\text{No-TCB}}$ variants reported in Section~\ref{sec:ablation} of the main paper. The comparison includes the Harmless Rate (HR) on malicious categories, as well as the False Rejection Rate (FRR) on both the general benign test set (GBT) and the TCB test set, shown in table~\ref{tab:app_TCB_test}.

\begin{table}[htbp]
\centering
\caption{$\delta_t$ and $\delta_v$'s FRR on the TCB test set, accompanying with other comparative data in the main text.}
\label{tab:app_TCB_test}
\tiny
\resizebox{0.98\textwidth}{!}{
\begin{tabular}{l ccc ccc ccc}
\toprule
 &
\multicolumn{3}{c}{\textbf{LLaMA-3.2-11B-Vision}} &
\multicolumn{3}{c}{\textbf{Qwen-VL-Chat}} &
\multicolumn{3}{c}{\textbf{LLaVA-1.5}} \\
\cmidrule(lr){2-4} \cmidrule(lr){5-7} \cmidrule(lr){8-10}
& HR & FRR (GBT) & FRR (TCB)
& HR & FRR (GBT) & FRR (TCB)
& HR & FRR (GBT) & FRR (TCB) \\
\midrule

ST-$\delta_v$ &  84.23 &7.75  &35.00  & 64.54 & 5.75 & 14.50 & 49.51 & 6.25 &20.5  \\

ST-$\delta_t$ & 81.89 &0.50  &38.00  & 65.56 & 1.75 & 16.50 & 51.98 & 1.50 & 4.5 \\

ST-$\delta_v^{\text{No-TCB}}$ &  58.75 &19.50  &93.00  & 40.15 & 23.00 & 98.75 & 31.50 & 17.50 &93.75  \\

ST-$\delta_t^{\text{No-TCB}}$ & 51.39 &15.00  &91.25  & 35.75 & 21.50 & 96.50 & 29.25 & 16.75 & 90.00 \\

\bottomrule
\end{tabular}
}
\end{table}

Compared to ST-$\delta_v^{\text{No-TCB}}$ and ST-$\delta_t^{\text{No-TCB}}$, incorporating the TCB set into training significantly reduces the over-rejection of benign queries that share similar textual structures with the SA set. This suggests that training security tensors on contrastive examples from the TCB set encourages them to rely more on visual information and reduces their dependence on textual patterns. However, incorporating the TCB set alone in training is not sufficient. As shown in our results, the FRR of $\delta_t$ and $\delta_v$ on the TCB test set remains considerably higher than their FRR on the general benign test set (GBT). This highlights the need for additional strategies beyond the TCB set to further mitigate text-pattern overfitting—a direction we leave for future work.


\end{document}